\documentclass{article}

\usepackage[preprint]{neurips_2026}

\usepackage[utf8]{inputenc}
\usepackage[T1]{fontenc}
\usepackage{hyperref}
\usepackage{url}
\usepackage{booktabs}
\usepackage{amsmath,amssymb,amsfonts,mathtools}
\usepackage{amsthm}
\usepackage{graphicx}
\usepackage{tabularx}
\usepackage{multirow}
\usepackage{enumitem}
\usepackage{algorithm}
\usepackage{algorithmic}
\usepackage{placeins}
\usepackage{tikz}
\usetikzlibrary{positioning,shapes,arrows.meta}

\newtheorem{theorem}{Theorem}

\newtheorem{remark}{Remark}

\newcommand{\E}{\mathbb{E}}

\title{Learning Adapter Rank via Symmetry Breaking}

\author{%
  Cooper Doyle \quad Andy Hu \quad Rebecca Chan \quad Anna Leontjeva \\
  Commonwealth Bank of Australia\\
}

\begin{document}

\maketitle


\begin{abstract}
Low-rank adaptation is effective partly because downstream updates lie in a low-dimensional subspace, but the latent rank coordinates of LoRA are not identifiable: any invertible reparameterization of the adapter factors leaves the weight update unchanged. We show that variational inference with a diagonal rank-wise posterior turns this non-identifiability into a useful inductive bias. By breaking LoRA's rotational gauge symmetry, the variational objective selects a preferred basis in rank space, enabling automatic relevance determination over rank directions. This yields Low-Rank Variational Dropout (LRVD), a Bayesian framework that performs inference directly in the low-rank adaptation space rather than the ambient weight space. As an instantiation, BayesLoRA jointly learns effective adapter rank and predictive uncertainty with only $\mathcal{O}(r)$ additional parameters. Empirically, BayesLoRA induces stable rank structure aligned with the dominant singular directions of learned updates, yields compact predictive calibration and matches or exceeds strong low-rank sparsification baselines at comparable training cost.
\end{abstract}

\section{Introduction}
\label{sec:intro}

Low-rank adaptation methods such as LoRA \citep{HuEtAl2022LoRA} have become a standard approach for fine-tuning large language models efficiently, with strong empirical evidence that task-relevant variation concentrates in a small number of directions \citep{Aghajanyan2021Subspace}. Yet two practical problems remain unsolved. First, choosing the right rank is non-trivial: fixed-rank methods can be wasteful or insufficient, and heuristic rank-allocation strategies introduce additional tuning burden. Second, and more subtly, the rank directions themselves are meaningless in standard LoRA: for any invertible $R \in \mathbb{R}^{r \times r}$, the factorization $BA = (BR)(R^{-1}A)$ leaves the weight update unchanged. This gauge freedom means individual rank indices carry no intrinsic identity, and any method that attaches uncertainty or relevance to specific rank directions must first deal with this non-identifiability.

Parameter-efficient fine-tuning also inherits the miscalibration problems of large models \citep{kadavath2022language, xiong2023can, kapoor2024large}: fine-tuned LLMs are prone to overconfidence. Existing uncertainty methods for PEFT modules \citep{balabanov2024uncertainty, wang2023lora, LaplaceLoRA, SWAGLoRA} treat rank selection and uncertainty quantification as separate problems, operating in the ambient weight space or applying post-hoc approximations, and so inherit architectural complexity and parameter count.

To address this, we formalise \textbf{Low-Rank Variational Dropout (LRVD)}, a variational dropout framework operating directly over latent rank directions. By tying posterior variance across each rank component, LRVD imposes a diagonal structure that breaks LoRA's rotational gauge symmetry, inducing structured sparsity through learned noise-to-signal ratios and enabling automatic rank selection. As a concrete instantiation, we introduce \textbf{BayesLoRA}, which learns the effective adapter rank during fine-tuning with only $\mathcal{O}(r)$ additional scalar parameters.

\begin{figure}[t]
    \centering
    \includegraphics[width=\linewidth]{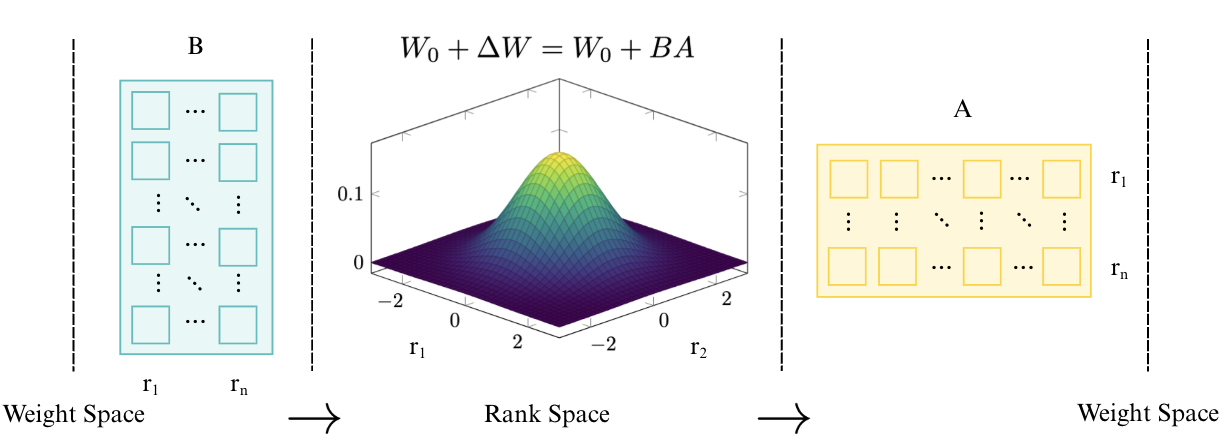}
    \caption{
    \textbf{Bayesian modeling in rank space.}
    Low-rank adaptation represents updates as $\Delta W = BA$, mapping parameters from weight space
    into a low-dimensional rank space.
    LRVD places uncertainty over this rank space rather than the full weight space, enabling structured
    uncertainty and rank-wise sparsification with minimal overhead.
    The surface is illustrative and does not represent a literal posterior density.
    }
    \label{fig:lrvd_rank_space}
\end{figure}

\section{Background and Related Work}
\label{sec:background}

\subsection{Bayesian and Variational Uncertainty in Neural Networks}

Bayesian approaches to uncertainty in neural networks include weight-space variational inference \citep{Blundell2015WeightUncertainty}, stochastic-gradient MCMC \citep{WellingTeh2011SGLD}, deep ensembles \citep{Lakshminarayanan2017DeepEnsembles}, SWAG-style covariance approximations \citep{Maddox2019SWAG}, and MC dropout \citep{GalGhahramani2016, KendallGal2017}. These methods differ substantially in cost and posterior expressiveness, but all parameterise uncertainty over the full weight space.

Our approach builds most directly on variational dropout 
\citep{Kingma2015LocalReparam, Molchanov2017VarDropSparsifies}, which parameterises each weight's posterior via a learned noise-to-signal ratio $\alpha_i$,
\begin{align}
    w_i \sim \mathcal{N}(\mu_i, \alpha_i \mu_i^2).
\end{align}
As $\alpha_i$ grows, the posterior variance dominates the mean and the parameter's functional contribution vanishes, inducing ARD-style sparsity through KL regularisation. LRVD applies this mechanism not to individual weights but to entire rank directions, which is what makes automatic rank selection possible.

Structured extensions \citep{Neklyudov2017SBP,Louizos2017BayesianCompression,LouizosWelling2017MNF} enable group-wise sparsity, but richer posterior structure typically remains prohibitively expensive at scale.

\subsection{Uncertainty in Low-Rank Adaptation}

Empirical results show that the intrinsic dimension of downstream adaptation is far smaller than the full parameter count \citep{Aghajanyan2021Subspace}, motivating LoRA \citep{HuEtAl2022LoRA} and rank-adaptive variants that learn or prune the effective rank \citep{Zhang2023adaloRA}.

Several recent methods combine low-rank adaptation with uncertainty estimation. LoRA-Ensemble \citep{Muhlematter2025LoRAEnsemble} improves calibration by averaging multiple independently trained adapters. Laplace-LoRA \citep{LaplaceLoRA} applies a \emph{post-hoc} Laplace approximation to LoRA parameters, requiring curvature estimation (e.g., Kronecker-factored structure) and linearised prediction. SWAG-LoRA \citep{SWAGLoRA} similarly constructs a post-hoc Gaussian approximation from the training trajectory via SWA/SWAG \citep{Maddox2019SWAG}.

In contrast, BloB \citep{BloB} learns a variational posterior over low-rank adaptations during fine-tuning, jointly updating posterior means and covariances throughout training. BloB achieves strong calibration but requires a separate variational optimiser, a carefully tuned KL annealing schedule, and approximately 1.5× the nominal LoRA parameter count. Projection-based posterior estimation \citep{Marszalek2025MinimalConfidence} models uncertainty inside a fixed low-dimensional subspace, but assumes the subspace rather than learning which rank directions are relevant.

\section{Low-Rank Variational Dropout (LRVD)}
\label{sec:lrvd_theory}

\textbf{Overview.} LRVD is a general framework for Bayesian inference that operates directly in rank space rather than weight space. \textbf{BayesLoRA} instantiates LRVD within the LoRA parameterisation, introducing only $\mathcal{O}(r)$ additional scalar parameters and yielding automatic rank selection as its primary output, with improved predictive calibration as a byproduct.

\subsection{Setup}

Consider a weight update constrained to a low-dimensional subspace,
\begin{align}
W = W_0 + \frac{\lambda}{r} BA
  = W_0 + \frac{\lambda}{r}\sum_{i=1}^r B_{\cdot i} A_{i\cdot}^\top,
\end{align}
where $W_0$ is frozen, $A \in \mathbb{R}^{r \times d_{\text{in}}}$,
$B \in \mathbb{R}^{d_{\text{out}} \times r}$, and $r \ll \min(d_{\text{in}},d_{\text{out}})$. LRVD treats this update as a structured random function whose stochasticity is confined to the adaptation subspace (Figure~\ref{fig:lrvd_rank_space}), preserving the determinism of the pretrained backbone.

Unlike weight-space Bayesian methods, LRVD does not posit a covariance model over $W$; randomness is confined to the span of the learned low-rank directions, keeping the backbone deterministic and inference cheap.

\subsection{Rank-Structured Posterior and Gauge Symmetry Breaking}
\label{sec:rank_structured_posterior}

We place a factorised Gaussian posterior over the adapter factors with
\emph{rank-tied} variances,

\begin{align}
q(A,B)
=
\prod_{i,j} \mathcal N(A_{ij};\,\mu_{A,ij},\,\alpha_i \mu_{A,ij}^2)\,
\prod_{k,i} \mathcal N(B_{ki};\,\mu_{B,ki},\,\alpha_i \mu_{B,ki}^2),
\end{align}

so all parameters belonging to rank $i$ share a single noise-to-signal ratio $\alpha_i$.
This yields a posterior with only $r$ uncertainty degrees of freedom, parameterised directly in rank space. By contrast with a mean-field posterior over the adapter factors, LRVD reduces the variational overhead by a factor of $d_{\mathrm{in}} + d_{\mathrm{out}}$.

\paragraph{Gauge symmetry breaking.}
LoRA has a latent gauge freedom: for any invertible $R\in\mathbb{R}^{r\times r}$,
$BA=(BR)(R^{-1}A)$, so rank indices carry no intrinsic identity.
Any method that attaches relevance to individual rank components must first break
this symmetry.
The diagonal (ARD-style) structure of our posterior cannot preserve the full gauge symmetry, and we exploit this: the variational objective selects a preferred basis in rank space and performs automatic relevance determination over rank directions simultaneously. The following theorem characterizes the residual symmetry group of the LRVD family.

\paragraph{Residual symmetry of the LRVD family.}
The symmetry-breaking effect of the rank-tied diagonal posterior can be stated precisely.

\begin{theorem}[Residual gauge symmetry of LRVD]
\label{thm:lrvd_residual_symmetry}
Let
\[
A \sim \mathcal{MN}(\mu_A, D, I), \qquad
B \sim \mathcal{MN}(\mu_B, I, D),
\]
where \(D = \mathrm{diag}(d_1,\dots,d_r) \succ 0\) is the shared rank-space covariance in the
variational family. Consider an invertible LoRA reparameterization
\[
A' = R^{-1}A, \qquad B' = BR, \qquad R \in GL(r),
\]
which leaves the mean update invariant since \(B'A' = BA\).

Suppose the pushed-forward distribution of \((A',B')\) also belongs to the same LRVD family with
a common diagonal rank covariance \(D'\), i.e.
\[
A' \sim \mathcal{MN}(R^{-1}\mu_A, D', I), \qquad
B' \sim \mathcal{MN}(\mu_B R, I, D'),
\]
for some diagonal \(D' \succ 0\). Then \(R\) must be orthogonal. Moreover,
\[
R^T D R = D'
\]
is diagonal, so \(R\) can only mix rank coordinates inside eigenspaces of \(D\). In particular,
if the diagonal entries of \(D\) are all distinct, then \(R\) is a signed permutation matrix.
\end{theorem}

\begin{proof}[Proof sketch]
Applying the matrix-normal covariance transformation under $A' = R^{-1}A$, $B' = BR$ gives rank covariances $R^{-1}DR^{-T}$ and $R^TDR$ respectively. Requiring both to equal the same diagonal $D'$ yields $RR^TDRR^T = D$, from which a positive-definite square root argument forces $RR^T = I$, so $R$ is orthogonal. Diagonality of $R^TDR$ then restricts $R$ to mix coordinates only within eigenspaces of $D$; when entries of $D$ are distinct, only signed permutations remain. Full proof in Appendix~\ref{app:proof}.
\end{proof}

\begin{remark}
The multiplicative variational-dropout posterior used in BayesLoRA retains an additional harmless rank-wise scaling gauge: for any diagonal invertible $S$, the transformation $A'=S^{-1}A, B'=BS$ preserves both the mean update and the multiplicative noise-to-signal ratios $\alpha_i$. This residual scaling symmetry does not affect rank relevance, since $\alpha_i$ and the pruning decision are invariant under it.
\end{remark}

\subsection{Variational Objective}

Training maximises the ELBO,
\begin{align}
\mathcal{L}
=
\mathbb{E}_{q(A,B)}\!\left[
  \log p\!\left(D \mid W_0 + \tfrac{\lambda}{r}BA\right)
\right]
+ \beta \sum_{i=1}^r \mathrm{KL}_i,
\end{align}
where $\beta$ controls regularisation strength.
Due to variance tying, the KL decomposes additively over rank components, directly penalising entire directions rather than individual weights. Each rank component uses the sparse variational dropout approximation of \citet{Molchanov2017VarDropSparsifies} to the KL contribution, up to an additive constant (constants and derivation in Appendix~\ref{app:kl}):
\begin{align}
\mathrm{KL}(\alpha)
=
k_1\,\sigma(k_2 + k_3\log\alpha)
- \tfrac{1}{2}\log(1+\alpha^{-1})
- k_1,
\end{align}
which is monotonically increasing in $\log\alpha$.

Since the KL contribution is monotonically increasing in $\log\alpha_i$, rank components with large $\alpha_i$ are progressively suppressed, yielding continuous automatic rank selection, while the likelihood keeps useful rank directions at low noise. At convergence, we define the effective rank as $r_\text{eff} = \sum_{i=1}^r \mathbf{1}[\log \alpha_i < \tau]$ and fix $\tau = 4$ throughout; performance is stable across a broad range (Figure~\ref{fig:tau_ablation}).

\subsection{Inference}

At test time we marginalise over the adapter subspace via $T$ Monte Carlo samples,
\begin{align}
\hat{p}(y\mid x)
\approx
\frac{1}{T}\sum_{t=1}^T p\!\left(y\mid x,\,W_0+\Delta W^{(t)}\right),\quad
\Delta W^{(t)}\sim q(A,B),
\end{align}
while the backbone remains deterministic; $T\in[4,16]$ suffices in practice.
To reduce gradient variance during training we apply local reparameterisation
\citep{Kingma2015LocalReparam}, moment-matching the adapter output in activation
space rather than sampling $A$ and $B$ directly (Appendix~\ref{app:derivations}).

\section{Rank Structure and Symmetry Breaking}
\label{sec:compression}

We first evaluate whether BayesLoRA selects compact ranks without sacrificing accuracy, then examine whether the selected rank structure is organised across modules and aligned with the intrinsic singular structure of the learned updates. To do this, we fine-tune DeBERTa-V3-base on the General Language Understanding Evaluation (GLUE) tasks \citep{wang2018glue} using BayesLoRA and compare against AdaLoRA and LoRA.

BayesLoRA begins with an over-complete low-rank adapter ($r=16$) and prunes rank components during training via automatic relevance determination. We then match AdaLoRA to BayesLoRA using an \emph{average rank budget}. Concretely, we compute BayesLoRA's average final \emph{effective rank} (defined as the mean of active ranks across all adapters) and run PEFT AdaLoRA with the nearest feasible integer budget (PEFT requires an integer budget). All results are averaged over three random seeds unless otherwise noted.

Table~\ref{tab:glue_slim} reports accuracy, final total rank, and training time. BayesLoRA matches or exceeds AdaLoRA across all tasks in comparable training time, with gains largest on CoLA and RTE — precisely the tasks where ARD compression is most aggressive. On tasks where little pruning occurs (MNLI, QNLI), performance is at parity, consistent with BayesLoRA recovering genuine low-dimensional task structure.

\begin{table*}[!htbp]
\centering
\scriptsize
\renewcommand{\arraystretch}{0.9}
\setlength{\tabcolsep}{3pt}
\caption{\textbf{GLUE benchmark (DeBERTa-V3 base).}
Best values per task are bolded. BayesLoRA matches or 
exceeds AdaLoRA across all tasks, with the largest gains 
on CoLA and RTE where compression is most aggressive.\\}
\resizebox{\textwidth}{!}{%
\begin{tabular}{llccccccc}
\toprule
Metric & Method & CoLA & MNLI & MRPC & QNLI & QQP & RTE & SST-2 \\
\midrule
\multirow{3}{*}{Acc ($\uparrow$)}
 & LoRA      & 0.858$_{.011}$ & 0.896$_{.000}$ & 0.832$_{.128}$ & 0.938$_{.001}$ & 0.900$_{.000}$ & 0.829$_{.021}$ & 0.952$_{.006}$ \\
 & AdaLoRA   & 0.874$_{.002}$ & 0.898$_{.001}$ & 0.900$_{.005}$ & 0.944$_{.001}$ & 0.897$_{.001}$ & 0.860$_{.015}$ & \textbf{0.957$_{.001}$} \\
 & BayesLoRA & \textbf{0.882$_{.005}$} & \textbf{0.899$_{.001}$} & \textbf{0.905$_{.003}$} & \textbf{0.944$_{.002}$} & \textbf{0.901$_{.000}$} & \textbf{0.871$_{.012}$} & 0.956$_{.002}$ \\
\midrule
\multirow{3}{*}{Rank ($\downarrow$)}
 & LoRA      & \textbf{10} & 16 & 2 & \textbf{15} & 16 & 2 & 12 \\
 & AdaLoRA   & \textbf{10} & 16 & 1.67$_{.58}$ & 15.33$_{.58}$ & 16 & 2.33$_{1.53}$ & 12.33$_{.58}$ \\
 & BayesLoRA & 10.13$_{.06}$ & \textbf{15.88$_{.01}$} & \textbf{1.63$_{.21}$} & 15.50$_{.06}$ & \textbf{15.62$_{.06}$} & \textbf{2.31$_{1.09}$} & \textbf{12.18$_{.54}$} \\
\midrule
\multirow{3}{*}{Time (min)}
 & LoRA      & \textbf{12.3$_{.03}$} & \textbf{51.2$_{.05}$} & \textbf{24.9$_{.02}$} & \textbf{44.2$_{.01}$} & \textbf{94.7$_{.15}$} & \textbf{39.5$_{.01}$} & \textbf{15.2$_{.00}$} \\
 & AdaLoRA   & 24.9$_{.09}$ & 66.1$_{.06}$ & 38.4$_{.03}$ & 59.6$_{1.05}$ & 116.2$_{.09}$ & 56.7$_{.03}$ & 26.5$_{.17}$ \\
 & BayesLoRA & 35.9$_{.20}$ & 69.3$_{.12}$ & 37.4$_{.32}$ & 62.2$_{.09}$ & 114.4$_{.17}$ & 53.4$_{.74}$ & 37.0$_{.45}$ \\
\bottomrule
\end{tabular}
}
\label{tab:glue_slim}
\end{table*}

\subsection{Proportional Pruning and Intrinsic Task Structure}
\label{sec:proportional_pruning}

Table~\ref{tab:rank_pct} shows final effective rank as a percentage of initial rank across $r_{\mathrm{init}} \in \{8, 16\}$. Despite doubling the initial basis, compression ratios remain surprisingly consistent across tasks — CoLA compresses to ~63\% at both ranks, RTE to ~13\%. This proportionality is a feature of LRVD: ARD latches onto intrinsic task structure, so the fraction of directions pruned reflects properties of the data, not the initial rank. The required KL scale $\beta$ decreases mildly with $r_{\mathrm{init}}$, consistent with the total KL penalty scaling with the number of rank components; we discuss this and the effect of $\beta$ on the accuracy--compression frontier in Appendix~\ref{app:beta_sweep}.

\begin{table}[h]
\centering
\caption{BayesLoRA compression vs initial rank (\%)}
\begin{tabular}{lrrrrrrr}
\toprule
\textbf{r$_{init}$} & \textbf{CoLA} & \textbf{MNLI} & \textbf{MRPC} & \textbf{QNLI} & \textbf{QQP} & \textbf{RTE} & \textbf{SST-2} \\
\midrule
8  & 63.31 & 98.55 & 7.47  & 91.84 & 85.65 & 12.27 & 63.60 \\
16 & 63.31 & 99.22 & 10.21 & 96.88 & 97.63 & 14.44 & 76.13 \\
\bottomrule
\end{tabular}
\label{tab:rank_pct}
\end{table}

As shown in Appendix~\ref{app:rank_selection_patterns}, BayesLoRA rank patterns often match AdaLoRA, frequently collapsing entire adapter modules to zero effective rank. This suggests that BayesLoRA behaves less like a smooth budget allocation scheme and more like rank-wise relevance determination with module-level sparsity.

Appendix~\ref{app:tau_ablation} shows the effect of the pruning threshold $\tau$ on accuracy and effective rank: accuracy is maintained over two orders of magnitude while the resulting compression changes smoothly. The consistency of these patterns across $r_{\mathrm{init}} 
\in \{8,16\}$ is documented in Table~\ref{tab:rank_pct}.

\subsection{Breaking the LoRA Gauge Symmetry}
\label{sec:symmetry}
Theorem~\ref{thm:lrvd_residual_symmetry} establishes that LRVD breaks GL(r) gauge symmetry in theory; here we verify that training produces a basis with meaningful structure in practice. Since LoRA factors are only identifiable up to invertible reparameterisations, any component-wise notion of rank relevance is only meaningful if variational training breaks this symmetry. If so, the resulting rank ordering should align with intrinsic structure in the learned update.

To measure this, we use two adapters (\textsc{SST-2} and \textsc{MNLI}) and for each we compute the singular value decomposition (SVD) of the mean update $\Delta W_\mu = BA$ and measure cumulative energy capture as ranks are added. We compare three orderings: (i) the optimal SVD ordering, (ii) BayesLoRA's ordering induced by increasing $\log \hat\alpha$, and (iii) random permutations of the same learned rank components, which preserve $\Delta W_\mu$ but destroy rank semantics.

\begin{figure}[t]
  \centering
  \includegraphics[width=\linewidth]{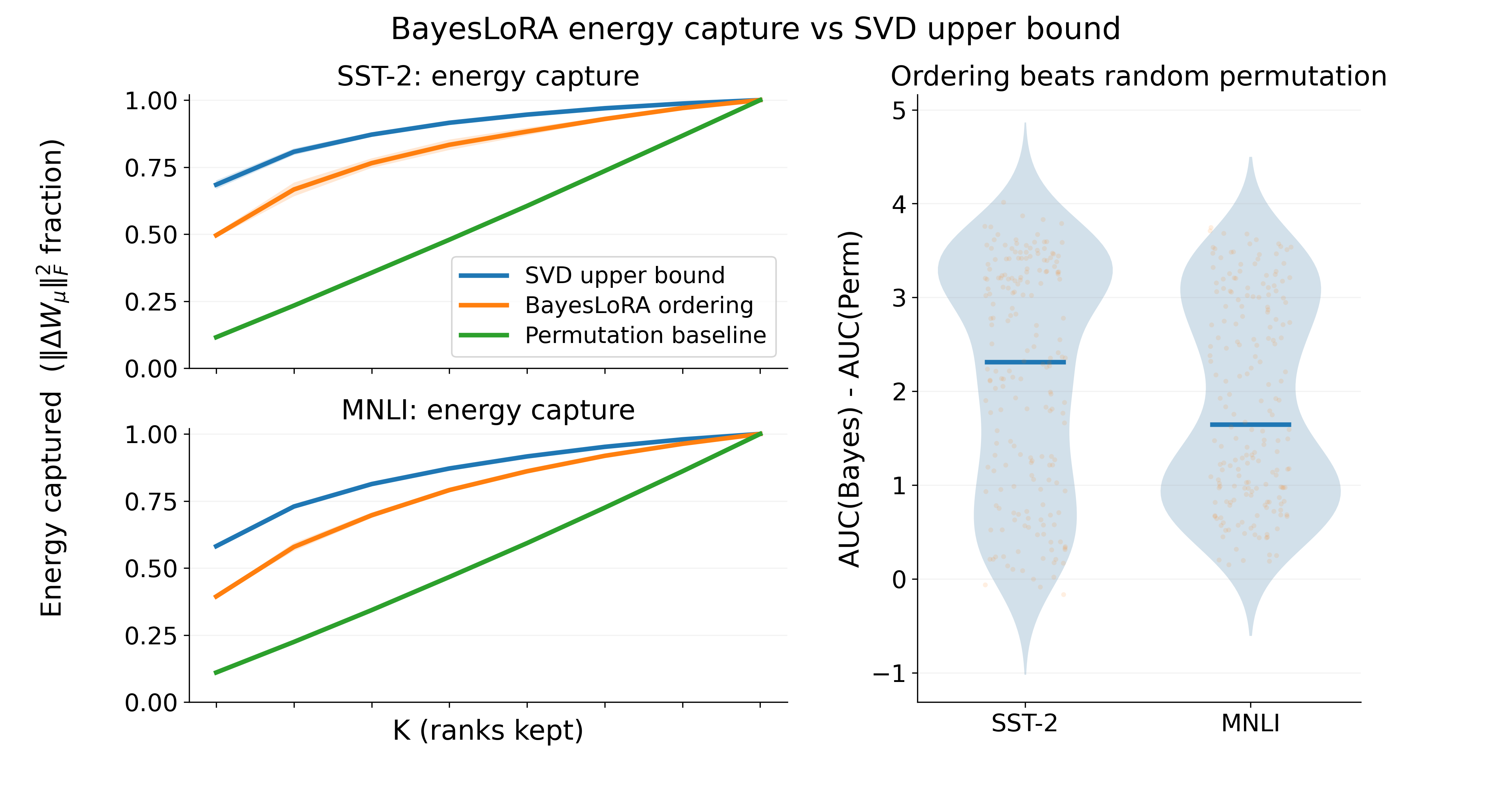}
  \caption{\textbf{Gauge symmetry breaking via Bayesian rank selection.}
  \emph{Left:} Cumulative energy capture as a function of retained rank, comparing the SVD upper bound (blue), BayesLoRA rank ordering (orange), and random permutations (green).
  \emph{Right:} Distribution of AUC improvements of BayesLoRA over random permutations across modules and seeds.
  BayesLoRA consistently recovers intrinsic structure while random orderings do not.}
  \label{fig:gauge_symmetry}
\end{figure}

Figure~\ref{fig:gauge_symmetry} shows that BayesLoRA's ordering consistently approaches the SVD upper bound substantially faster than random permutations, with low variance across seeds. Simultaneously, Appendix~\ref{app:beta_sweep} shows that across tasks, seeds and regularisation strengths, BayesLoRA consistently selects predictable ranks. Together, these results show that the rank-wise relevance scores learned by BayesLoRA are not merely pruning heuristics, but induce a stable ordering aligned with the intrinsic structure of the learned update.

\section{Downstream Uncertainty Effects}
\label{sec:downstream}

We now ask whether the learned rank-space posterior captures functionally relevant uncertainty. We evaluate accuracy and uncertainty across six reasoning and language-understanding benchmarks: \textsc{ARC-Challenge}, \textsc{ARC-Easy}, \textsc{WinoGrande-S}, \textsc{WinoGrande-M}, \textsc{OpenBookQA}, and \textsc{BoolQ}. All reasoning experiments use a Llama~2~7B backbone (full hyperparameters in Appendix~\ref{app:reasoning_bench}).

All models are fine-tuned for 5000 steps, with evaluation every 100 steps. We report final Accuracy, Expected Calibration Error (ECE), and Negative Log-Likelihood (NLL), including deterministic posterior-mean inference ($k=0$). All stochastic methods use Monte Carlo posterior marginalisation with $k = 10$. Appendix~\ref{app:mc_sample_sweep} provides an inference-time sample sweep showing the corresponding accuracy--calibration--latency trade-off.

\subsection{Rank-Space Uncertainty vs.\ Classical Baselines}

Table~\ref{tab:reasoning_main} compares BayesLoRA against standard uncertainty baselines that share its computational profile: maximum-likelihood (MLE), MAP regularisation, MC dropout~\citep{GalGhahramani2016}, deep ensembles~\citep{Lakshminarayanan2017DeepEnsembles}, and Bayes by Backprop (BBB)~\citep{Blundell2015WeightUncertainty}. We additionally include $\mathrm{BayesLoRA}_{r=8}$ (BayesLoRA with a fixed rank and no pruning) to disentangle the effect of learned low-rank uncertainty from the effect of rank selection.

As shown in Table~\ref{tab:reasoning_main}, BayesLoRA substantially improves calibration over most classical uncertainty baselines despite an improper prior. BBB performs variational inference over the same parameter space with more than twice the parameters and a comparable prior, yet produces worse calibration. This suggests that the calibration benefit is not just a consequence of averaging but of how uncertainty is parametrised.

\begin{table*}[t]
\centering
\caption{Reasoning benchmark results (LLaMA-2-7B), evaluated at 5000 steps. Mean$_{\mathrm{std}}$ over 3 seeds. Higher is better for Accuracy; lower is better for ECE, NLL, and Params. \textbf{Bold} indicates best within each block. Citations: BBB~\citep{Blundell2015WeightUncertainty}, MCD~\citep{GalGhahramani2016}, ENS~\citep{Lakshminarayanan2017DeepEnsembles}.}
\label{tab:reasoning_main}
\vspace{2pt}
\setlength{\tabcolsep}{2.5pt}
\scriptsize
\begin{tabular}{llc*{6}{c}}
\toprule
Metric & Method & Params (M)$\downarrow$ & WG-S & ARC-C & ARC-E & WG-M & OBQA & BoolQ \\
\midrule

\multirow{11}{*}{Acc.\,$\uparrow$}
& LoRA & 4.48 & 68.99$_{0.58}$ & 69.10$_{2.84}$ & 85.65$_{0.92}$ & 74.53$_{0.66}$ & 81.52$_{0.25}$ & 86.53$_{0.28}$ \\
& MCD & 4.48 & 69.46$_{0.62}$ & 68.69$_{1.30}$ & 86.21$_{0.46}$ & \textbf{76.45$_{0.04}$} & 81.72$_{0.10}$ & 87.29$_{0.13}$ \\
& ENS & 44.80 & 69.57$_{0.66}$ & 66.20$_{2.01}$ & 84.40$_{0.81}$ & 75.32$_{0.21}$ & 81.38$_{0.91}$ & 87.09$_{0.11}$ \\
& BBB & 9.98 & 56.54$_{7.87}$ & 68.13$_{1.27}$ & 85.86$_{0.74}$ & 73.63$_{2.44}$ & 82.06$_{0.59}$ & 87.21$_{0.22}$ \\
\cmidrule(lr){2-9}
& BayesLoRA$_{r=8}$ ($k=0$) & 4.48 & 69.64$_{2.07}$ & 68.68$_{1.69}$ & \textbf{86.96$_{0.10}$} & 75.45$_{0.44}$ & \textbf{82.80$_{1.00}$} & 87.42$_{0.50}$ \\
& BayesLoRA$_{r=8}$ ($k=10$) & 4.48 & 68.53$_{1.75}$ & 68.90$_{0.39}$ & 86.14$_{0.53}$ & 74.66$_{0.65}$ & 81.27$_{0.90}$ & 87.14$_{0.24}$ \\
& BayesLoRA ($k=0$) & \textbf{2.40--3.87} & \textbf{70.43$_{0.87}$} & 68.37$_{2.04}$ & 86.73$_{0.37}$ & 75.94$_{0.14}$ & 82.07$_{0.50}$ & 87.54$_{0.11}$ \\
& BayesLoRA ($k=10$) & \textbf{2.40--3.87} & 69.40$_{0.92}$ & \textbf{69.46$_{2.61}$} & 86.32$_{0.61}$ & 75.22$_{1.11}$ & 81.40$_{0.53}$ & \textbf{87.60$_{0.15}$} \\
\midrule

\multirow{11}{*}{ECE\,$\downarrow$}
& LoRA & 4.48 & 29.83$_{0.58}$ & 29.00$_{1.97}$ & 13.12$_{1.39}$ & 20.62$_{0.74}$ & 12.55$_{0.46}$ & 3.18$_{0.09}$ \\
& MCD & 4.48 & 27.98$_{0.44}$ & 27.53$_{0.80}$ & 12.20$_{0.56}$ & 19.55$_{0.47}$ & 13.10$_{0.11}$ & 3.46$_{0.16}$ \\
& ENS & 44.80 & 28.52$_{0.55}$ & 29.16$_{2.37}$ & 12.57$_{0.58}$ & 20.86$_{0.43}$ & 15.34$_{0.27}$ & 9.61$_{0.24}$ \\
& BBB & 9.98 & 21.81$_{12.95}$ & 26.23$_{1.47}$ & 12.28$_{0.58}$ & 15.76$_{4.71}$ & 11.38$_{1.07}$ & 3.74$_{0.10}$ \\
\cmidrule(lr){2-9}
& BayesLoRA$_{r=8}$ ($k=0$) & 4.48 & 29.82$_{2.11}$ & 30.45$_{1.88}$ & 12.36$_{0.26}$ & 21.87$_{0.48}$ & 14.89$_{0.79}$ & 10.11$_{0.25}$ \\
& BayesLoRA$_{r=8}$ ($k=10$) & 4.48 & 21.09$_{1.26}$ & \textbf{17.47$_{0.38}$} & \textbf{7.70$_{0.43}$} & 14.97$_{0.59}$ & 10.01$_{0.47}$ & 7.34$_{0.06}$ \\
& BayesLoRA ($k=0$) & \textbf{2.40--3.87} & 29.26$_{1.06}$ & 30.73$_{1.92}$ & 12.79$_{0.27}$ & 22.18$_{0.23}$ & 15.37$_{0.62}$ & 9.73$_{0.18}$ \\
& BayesLoRA ($k=10$) & \textbf{2.40--3.87} & \textbf{20.62$_{0.71}$} & 17.49$_{2.40}$ & 7.99$_{0.58}$ & \textbf{14.46$_{0.83}$} & \textbf{9.38$_{0.61}$} & \textbf{6.87$_{0.43}$} \\
\midrule

\multirow{11}{*}{NLL\,$\downarrow$}
& LoRA & 4.48 & 3.17$_{0.37}$ & 2.85$_{0.27}$ & 1.17$_{0.13}$ & 0.95$_{0.07}$ & 0.73$_{0.03}$ & 0.32$_{0.00}$ \\
& MCD & 4.48 & 2.79$_{0.53}$ & 2.67$_{0.15}$ & 1.00$_{0.14}$ & 1.02$_{0.03}$ & 0.77$_{0.03}$ & \textbf{0.31$_{0.00}$} \\
& ENS & 44.80 & 2.71$_{0.08}$ & 2.46$_{0.22}$ & 0.82$_{0.03}$ & 1.25$_{0.03}$ & 1.06$_{0.04}$ & 0.57$_{0.02}$ \\
& BBB & 9.98 & 1.40$_{0.55}$ & 2.23$_{0.04}$ & 0.91$_{0.06}$ & 0.84$_{0.15}$ & 0.66$_{0.05}$ & \textbf{0.31$_{0.00}$} \\
\cmidrule(lr){2-9}
& BayesLoRA$_{r=8}$ ($k=0$) & 4.48 & 2.69$_{0.41}$ & 2.93$_{0.15}$ & 0.94$_{0.05}$ & 1.13$_{0.03}$ & 0.97$_{0.01}$ & 0.50$_{0.01}$ \\
& BayesLoRA$_{r=8}$ ($k=10$) & 4.48 & \textbf{1.37$_{0.07}$} & \textbf{1.40$_{0.06}$} & \textbf{0.55$_{0.03}$} & 0.75$_{0.02}$ & 0.64$_{0.03}$ & 0.38$_{0.00}$ \\
& BayesLoRA ($k=0$) & \textbf{2.40--3.87} & 2.78$_{0.33}$ & 2.99$_{0.05}$ & 0.91$_{0.02}$ & 1.11$_{0.01}$ & 0.95$_{0.08}$ & 0.49$_{0.00}$ \\
& BayesLoRA ($k=10$) & \textbf{2.40--3.87} & 1.45$_{0.10}$ & 1.43$_{0.04}$ & \textbf{0.55$_{0.01}$} & \textbf{0.74$_{0.02}$} & \textbf{0.62$_{0.03}$} & 0.38$_{0.00}$ \\
\bottomrule
\end{tabular}
\end{table*}

\subsection{Matched-Budget Comparison with Calibration-Focused Methods}
\label{sec:head_to_head}

We next compare BayesLoRA against methods designed specifically for calibration: BloB~\citep{BloB}, which learns a full variational posterior over LoRA parameters with a separate optimiser and KL annealing schedule ($\sim$6.72M params, $\sim$$1.5\times$ nominal LoRA count), and Laplace-LoRA~\citep{LaplaceLoRA}, which applies post-hoc Kronecker-factored curvature estimation. Neither performs capacity control. To match parameter budgets, we initialise BayesLoRA with expanded per-task rank and reduce $\beta$ by $1/3$ so that the final count after pruning is comparable to BloB's fixed budget.

As shown in Table~\ref{tab:head_to_head}, BayesLoRA achieves higher accuracy across almost all tasks while matching or narrowing the calibration gap. BloB retains an edge on ECE and NLL, reflecting its richer variational machinery, but BayesLoRA achieves competitive calibration without auxiliary optimisers.

\begin{table*}[t]
\centering
\caption{Matched-budget comparison with calibration-focused methods (LLaMA-2-7B). BayesLoRA is initialised with expanded per-task rank so that the final parameter count after ARD pruning is comparable to BloB's fixed budget ($\sim$6.72M). \textbf{Bold} indicates best per metric. BayesLoRA achieves higher accuracy across almost all tasks while BloB retains stronger calibration, exposing a clear accuracy--calibration trade-off.\\}
\label{tab:head_to_head}
\vspace{2pt}
\setlength{\tabcolsep}{3pt}
\scriptsize
\begin{tabular}{llc*{6}{c}}
\toprule
Metric & Method & Params (M) & WG-S & ARC-C & ARC-E & WG-M & OBQA & BoolQ \\
\midrule
\multirow{3}{*}{Acc.\,$\uparrow$}
& LAP & 4.48 & 69.20$_{1.50}$ & 66.78$_{0.69}$ & 80.05$_{0.22}$ & 75.55$_{0.36}$ & 82.12$_{0.67}$ & 86.95$_{0.09}$ \\
& BloB ($k\!=\!10$) & 6.72 & \textbf{69.07$_{0.34}$} & 68.81$_{1.09}$ & 85.56$_{0.35}$ & 73.69$_{0.17}$ & 81.52$_{0.74}$ & 86.99$_{0.24}$ \\
& BayesLoRA$^{\dagger}$ ($k\!=\!10$) & $\sim$6.72 & 68.75$_{1.90}$ & \textbf{69.59$_{1.37}$} & \textbf{86.00$_{0.51}$} & \textbf{77.53$_{1.36}$} & \textbf{81.80$_{0.83}$} & \textbf{87.61$_{0.43}$} \\
\midrule
\multirow{3}{*}{ECE\,$\downarrow$}
& LAP & 4.48 & \textbf{4.15$_{1.12}$} & 16.25$_{2.61}$ & 33.29$_{0.57}$ & 7.40$_{0.27}$ & 8.70$_{1.77}$ & \textbf{1.30$_{0.33}$} \\
& BloB ($k\!=\!10$) & 6.72 & 9.35$_{1.37}$ & \textbf{9.59$_{1.88}$} & \textbf{3.64$_{0.53}$} & \textbf{3.01$_{0.12}$} & \textbf{3.77$_{1.47}$} & 1.41$_{0.19}$ \\
& BayesLoRA$^{\dagger}$ ($k\!=\!10$) & $\sim$6.72 & 24.00$_{1.78}$ & 19.40$_{1.50}$ & 9.30$_{0.45}$ & 14.86$_{1.09}$ & 10.93$_{0.59}$ & 7.28$_{0.05}$ \\
\bottomrule
\end{tabular}
\end{table*}

\section{Conclusion}

We introduced Low-Rank Variational Dropout (LRVD), a framework that places Bayesian inference directly in the rank space of low-rank adaptations rather than the ambient weight space. By showing that a diagonal variational posterior breaks LoRA's GL$(r)$ gauge symmetry (Theorem~1), we established that rank-wise relevance scores are generically identifiable --- turning a non-identifiability that previous methods ignored into a useful inductive bias.

BayesLoRA, a concrete instantiation of LRVD, adds only $\mathcal{O}(r)$ scalar parameters to standard LoRA and requires no auxiliary optimiser, no post-hoc curvature estimation, and no KL annealing schedule. Despite this simplicity, it simultaneously (i)~learns effective adapter rank and intrinsic task structure, often matching or beating AdaLoRA; (ii)~recovers a rank ordering aligned with the dominant singular directions of the learned update; and (iii)~substantially improves predictive calibration over classical uncertainty baselines including MC~dropout, deep ensembles, and Bayes by Backprop. In matched-budget comparisons against BloB, BayesLoRA achieves consistently higher accuracy while narrowing the calibration gap, exposing a clear accuracy--calibration trade-off between rank-space and weight-space variational methods.

\paragraph{Limitations.} BayesLoRA's calibration still lags behind methods with richer variational machinery such as BloB, particularly on tasks where the gap between rank-space and weight-space posteriors is most pronounced. The variational-dropout surrogate also inherits an improper log-uniform prior, which may limit the informativeness of the posterior in low-data regimes.

\paragraph{Future work.} A natural extension is to replace the variational-dropout surrogate with a hierarchical ARD prior over rank-specific scales, preserving the symmetry-breaking structure of LRVD while yielding a proper posterior that may help close the calibration gap. More broadly, LRVD applies wherever adaptation is intrinsically low-dimensional: low-rank attention, spectral compression, and prompt subspaces are natural candidates. Extending the framework to structured covariance models that capture inter-rank correlations without sacrificing the $\mathcal{O}(r)$ parameter overhead is another promising direction.

\FloatBarrier
\newpage
\bibliographystyle{plainnat}
\bibliography{references}

\newpage
\appendix

\section{Proof of Theorem~\ref{thm:lrvd_residual_symmetry}}
\label{app:proof}

\begin{proof}
Under the reparameterization \(A' = R^{-1}A\), \(B' = BR\), standard matrix-normal covariance
transformations give
\[
A' \sim \mathcal{MN}(R^{-1}\mu_A,\; R^{-1} D R^{-T},\; I),
\]
\[
B' \sim \mathcal{MN}(\mu_B R,\; I,\; R^T D R).
\]
If both transformed rank covariances are equal to the same diagonal matrix \(D'\), then
\[
R^{-1} D R^{-T} = D', \qquad R^T D R = D'.
\]
Combining these yields
\[
R R^T D R R^T = D.
\]
Let \(M = R R^T\). Since \(R\) is invertible, \(M \succ 0\), and the above becomes
\[
M D M = D.
\]
Now define
\[
X = D^{1/2} M D^{1/2}.
\]
Then \(X \succ 0\), and
\[
X^2
= D^{1/2} M D M D^{1/2}
= D^{1/2} D D^{1/2}
= D^2.
\]
Because \(X\) is positive definite and \(D \succ 0\), uniqueness of the positive-definite square
root implies \(X = D\). Therefore
\[
D^{1/2} M D^{1/2} = D,
\]
hence \(M = I\), i.e. \(R R^T = I\). Thus \(R\) is orthogonal.

It remains to characterize which orthogonal matrices preserve diagonality of \(R^T D R\). Since
\(D\) is already diagonal, an orthogonal \(R\) can only mix coordinates within eigenspaces
corresponding to equal diagonal entries of \(D\). Therefore the residual symmetry group is the
product of orthogonal groups on the repeated-eigenvalue blocks of \(D\). When the entries of
\(D\) are all distinct, these blocks are one-dimensional, so the only possibilities are signed
permutations.
\end{proof}

\section{Additional Method Details}
\label{app:derivations}

\subsection{Rank-Structured Variational Family}

BayesLoRA employs a structured mean-field variational posterior over the low-rank adapter factors.

Recall the parameterisation
\begin{align}
W
=
W_0
+
\frac{\lambda}{r}
\sum_{i=1}^{r}
B_{\cdot i} A_{i\cdot}.
\end{align}

The variational posterior factorises elementwise,
\begin{align}
q(A,B)
=
\prod_{i,j}
\mathcal N(A_{ij}; \mu_{A,ij}, \alpha_i \mu_{A,ij}^2)
\prod_{k,i}
\mathcal N(B_{ki}; \mu_{B,ki}, \alpha_i \mu_{B,ki}^2),
\end{align}
where all parameters associated with rank $i$ share a single noise-to-signal ratio $\alpha_i$.
This induces a structured posterior with only $r$ uncertainty degrees of freedom.

\subsection{Posterior Mean}

Taking expectation under $q(A,B)$ yields
\begin{align}
\E_q[W]
=
W_0
+
\frac{\lambda}{r}
\sum_{i=1}^{r}
\boldsymbol\mu_{B,i}
\boldsymbol\mu_{A,i}^\top,
\end{align}
which is itself low-rank and lies in the span of the learned rank directions. This expression corresponds to the deterministic adapter used at test time when stochastic sampling is disabled.

\subsection{Stochastic forward pass via local reparameterisation}

We propagate uncertainty without explicitly sampling $A$ and $B$.

Let $x \in \mathbb{R}^{d_{\mathrm{in}}}$ and define
\begin{align}
s = xA^\top.
\end{align}

Under the rank-tied variational family, the mean is
\begin{align}
m_s = x\mu_A^\top,
\end{align}
and the variance becomes
\begin{align}
v_s = (x \odot x)(\alpha \odot \mu_A^2)^\top,
\end{align}
where $\alpha \odot \mu_A^2$ applies the rank-wise scaling.

The adapter output $y = sB^\top$ has mean
\begin{align}
m_y = m_s \mu_B^\top,
\end{align}
and diagonal variance
\begin{align}
v_y =
v_s(\mu_B \odot \mu_B)^\top
+
(m_s \odot m_s)(\alpha \odot \mu_B^2)^\top.
\end{align}

We sample
\begin{align}
y \approx m_y + \epsilon \odot \sqrt{v_y + \varepsilon}, \quad \epsilon \sim \mathcal{N}(0,I).
\end{align}

\section{KL Approximation and Variational Dropout Details}
\label{app:kl}

\subsection{Elementwise Noise-to-Signal Ratio}

In this formulation, uncertainty is fully governed by the rank-level parameter $\alpha_i$.
No separate variance parameter is required.

For all elements in rank $i$, the noise-to-signal ratio is constant:
\begin{align}
\alpha = \alpha_i.
\end{align}

Rank relevance is assessed by aggregating $\log \alpha_i$ across parameters belonging to each rank component.

\subsection{KL Approximation}

The KL divergence between the variational posterior and a log-uniform prior follows
\citet{Molchanov2017VarDropSparsifies}:
\begin{align}
\mathrm{KL}(\alpha)
=
k_1\,\sigma(k_2 + k_3 \log \alpha)
-
\tfrac{1}{2}\log(1+\alpha^{-1})
-
k_1,
\end{align}
which is applied per rank component and summed over $i$.

\subsection{Constants}

We use the constants reported by \citet{Molchanov2017VarDropSparsifies}:
\begin{center}
\begin{tabular}{c c c}
\toprule
Constant & Value & Source \\
\midrule
$k_1$ & $0.63576$ & \citet{Molchanov2017VarDropSparsifies} \\
$k_2$ & $1.87320$ & \citet{Molchanov2017VarDropSparsifies} \\
$k_3$ & $1.48695$ & \citet{Molchanov2017VarDropSparsifies} \\
\bottomrule
\end{tabular}
\end{center}

\subsection{Rank Pruning Criterion}

A rank component is considered inactive when its aggregated noise-to-signal ratio exceeds a threshold.
Specifically, we prune rank $i$ if
\begin{align}
\log \hat\alpha_i > \tau,
\end{align}
where $\log \hat\alpha_i$ is the rank-level statistic defined in Section~\ref{sec:lrvd_theory}.

\section{GLUE Experiment Hyperparameters}
\label{app:glue_hparams}

We evaluate BayesLoRA and AdaLoRA on the GLUE benchmark using a DeBERTa-v3-base backbone. All results are averaged over three random seeds. Table~\ref{tab:glue_hparams} lists the hyperparameters used across all tasks.

\begin{table}[h]
\centering
\small
\caption{Hyperparameters used for BayesLoRA experiments.}
\begin{tabular}{l c}
\toprule
\textbf{Setting} & \textbf{Value} \\
\midrule
Max sequence length & 128 \\
Batch size & 32 \\
Optimizer & AdamW \\
Warmup ratio & 0.06 \\
Weight decay & $1\times10^{-2}$ \\
\midrule
Adapter learning rate (A,B) & $1\times10^{-3}$ \\
Classifier head learning rate & $1\times10^{-2}$ \\
LoRA initial rank $r_{\text{init}}$ & 8 \\
LoRA scale $\lambda$ & 16 \\
\midrule
Bayesian KL weight $\beta$ & $1\times10^{-6}$ \\
Log $\alpha$ threshold $\tau$ & 4.0 \\
Monte Carlo samples (inference) & $T = 4$ \\
Initial log $\sigma$ & $-8.0$ \\
Log $\sigma^2$ clamp range & $[-20,\;2]$ \\

\bottomrule
\end{tabular}
\label{tab:glue_hparams}
\end{table}

\begin{figure*}[t]
    \centering
    \includegraphics[width=\textwidth]{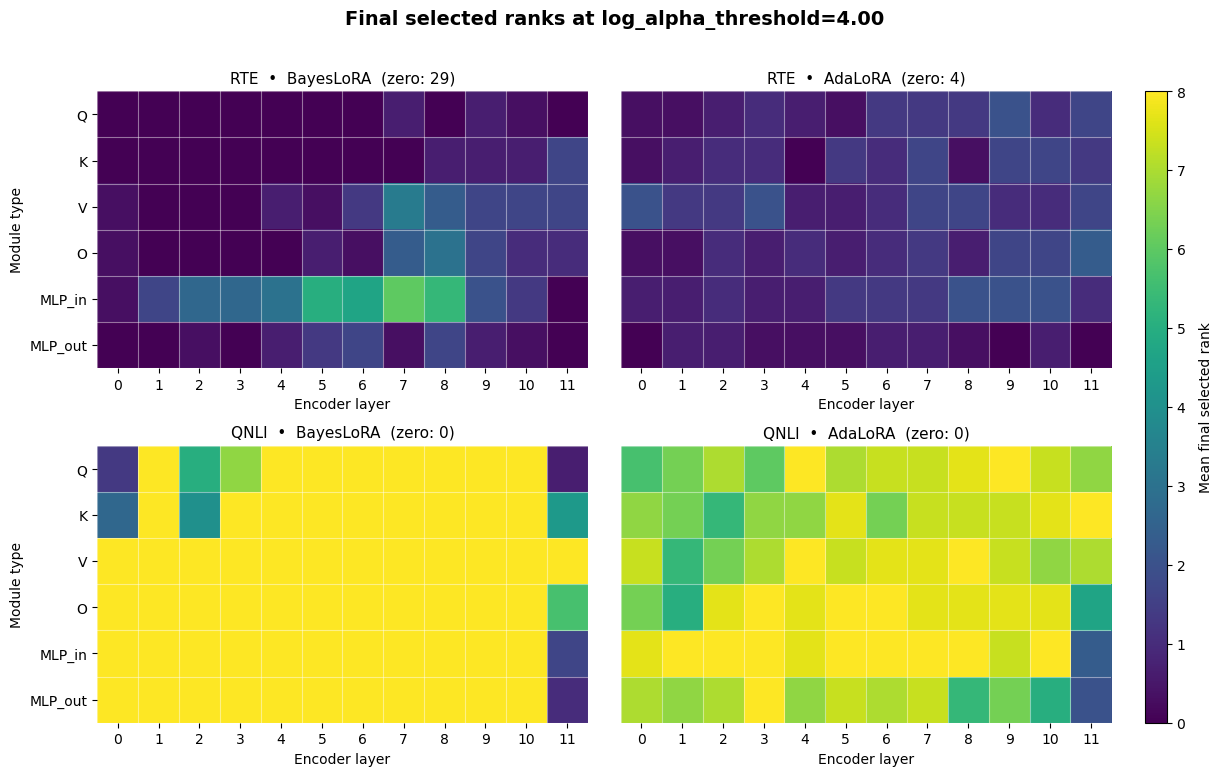}
    \caption{
    Final selected ranks at $\tau = 4.0$ for BayesLoRA and AdaLoRA on representative DeBERTa-v3 tasks. BayesLoRA exhibits a more selective final rank pattern, often driving entire adapter modules to zero effective rank, whereas AdaLoRA tends to retain a smoother nonzero allocation across layers and modules. The contrast is especially pronounced on RTE, where BayesLoRA eliminates many more modules entirely despite operating at a comparable overall budget.
    }
    \label{fig:rank_selection_patterns}
\end{figure*}

\section{Rank Selection Dynamics}
\label{app:rank_selection_dynamics}

\begin{figure*}[t]
    \centering
    \includegraphics[width=\textwidth]{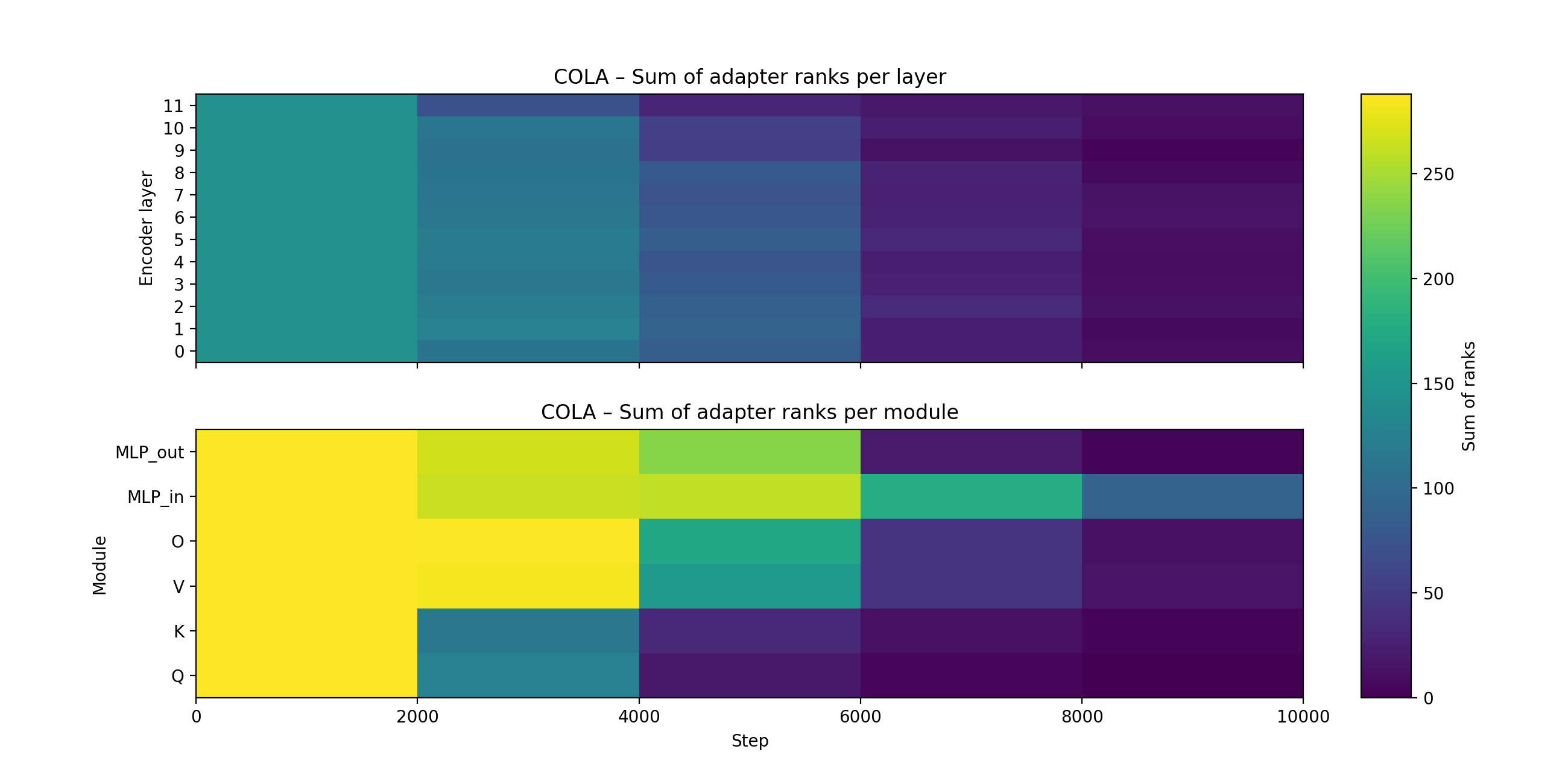}
    \caption{
    \textbf{Rank dynamics during fine-tuning on CoLA (DeBERTa-v3-base).}
    Heatmaps show the sum of active adapter ranks over training steps.
    \emph{Top:} distribution of effective rank across encoder layers.
    \emph{Bottom:} distribution across model modules (attention projections and MLP).
    BayesLoRA progressively concentrates capacity into a small subset of layers and modules,
    illustrating structured, data-driven rank pruning via rank-wise variational dropout.
    }
    \label{fig:rank_dynamics}
\end{figure*}

Figure~\ref{fig:rank_dynamics} visualises the evolution of active adapter ranks across layers and modules. Pruning is highly structured: many attention projections in higher layers are removed early, while a small number of MLP and value-projection subspaces remain active. This suggests that task-relevant signal concentrates in a limited number of directions, which BayesLoRA identifies without manual architectural choices.

\section{Final Rank Selection Patterns}
\label{app:rank_selection_patterns}

Figure~\ref{fig:rank_selection_patterns} compares the final selected rank patterns learned by BayesLoRA and AdaLoRA. Although both methods adapt rank across layers and modules, they occupy different points in the design space: AdaLoRA primarily redistributes a budget smoothly, whereas BayesLoRA tends to make sharper module-level sparsity, causing some adapters to collapse entirely to zero effective rank.

\begin{figure*}[t]
    \centering
    \includegraphics[width=\textwidth]{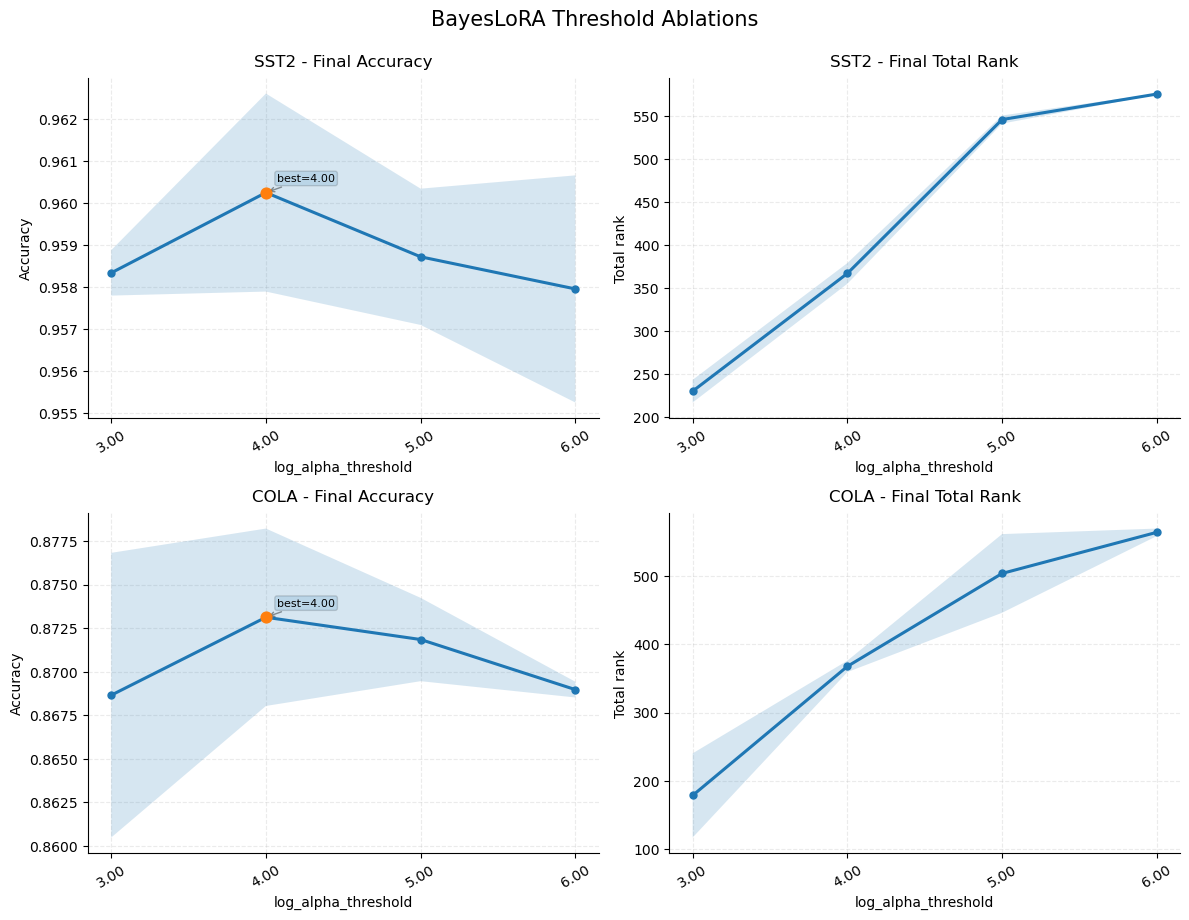}
    \caption{
    Effect of pruning threshold $\tau$ on final accuracy and total rank for representative GLUE tasks. Increasing $\tau$ makes pruning less aggressive and therefore increases the retained rank. Across a broad range of thresholds, accuracy remains relatively stable while compression changes smoothly, indicating that $\tau$ provides a simple and interpretable knob for the accuracy--compression trade-off.
    }
    \label{fig:tau_ablation}
\end{figure*}

\section{Effect of the Pruning Threshold}
\label{app:tau_ablation}

Figure~\ref{fig:tau_ablation} shows the effect of varying the pruning threshold $\tau$. As expected, larger values of $\tau$ retain more rank, since fewer directions are classified as inactive. At the same time, accuracy remains comparatively stable over a broad range, suggesting that BayesLoRA exposes a practical one-parameter trade-off between compression and performance.

\section{Effect of \texorpdfstring{$\beta$}{beta} on Accuracy and Rank}
\label{app:beta_sweep}

\begin{figure}[h]
    \centering
    \includegraphics[width=\linewidth]{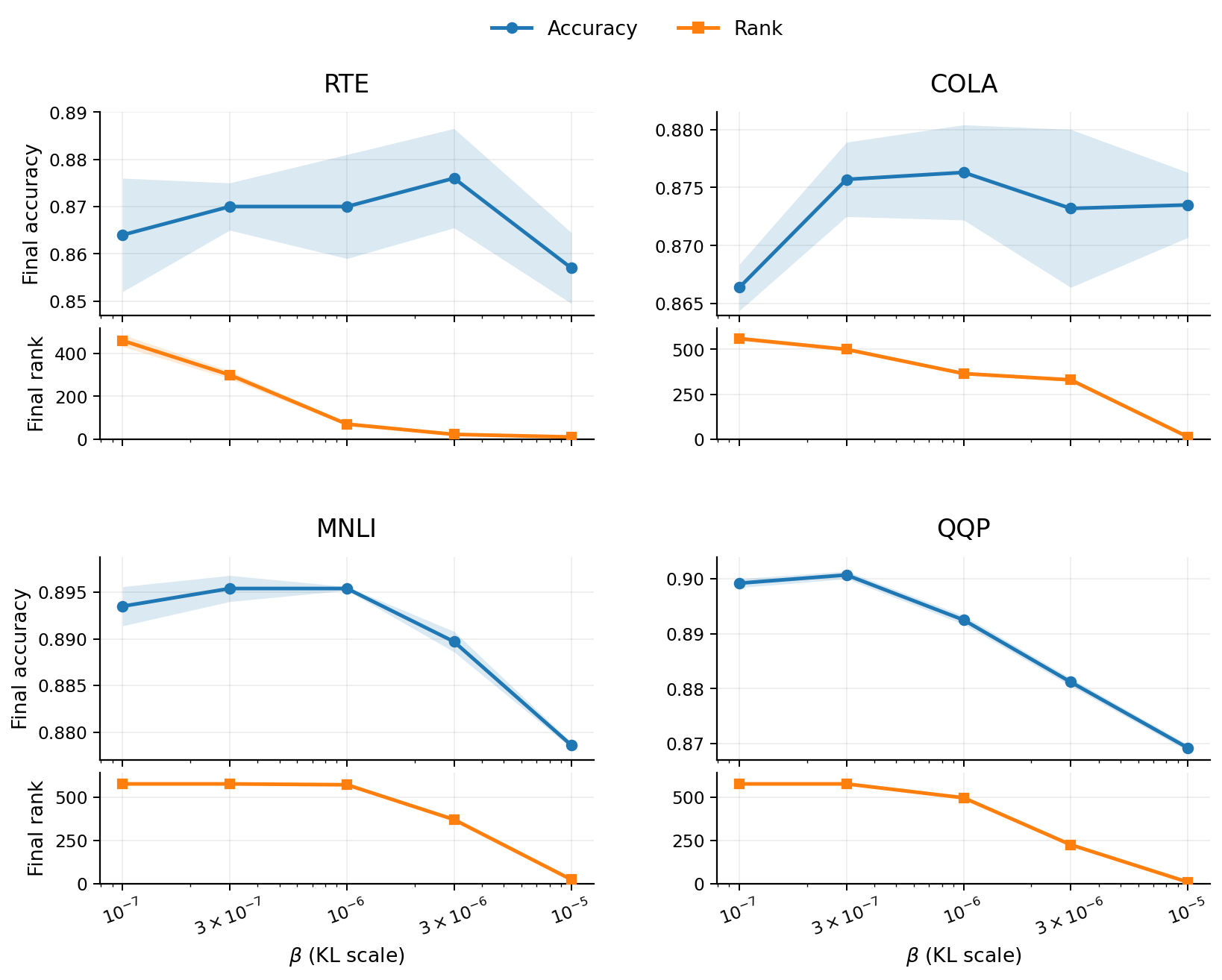}
    \caption{Effect of KL scale $\beta$ on final accuracy (blue, top) and effective rank
    (orange, bottom) for four GLUE tasks. Smaller datasets (RTE, CoLA) benefit from
    stronger regularisation, with accuracy peaking at moderate-to-high $\beta$ alongside
    aggressive rank compression. Larger datasets (MNLI, QQP) are sensitive to
    over-regularisation: accuracy is stable at low $\beta$ but degrades sharply once the
    KL penalty compresses rank substantially. Error bars show standard deviation over three
    seeds.}
    \label{fig:beta_sweep}
\end{figure}

Figure~\ref{fig:beta_sweep} shows final accuracy and effective rank as a function of the KL scale $\beta$ across four GLUE tasks, spanning a wide range of dataset sizes.

The pattern is consistent with the role of dataset size in determining useful adaptation capacity. On smaller datasets (RTE, CoLA), accuracy peaks at moderate-to-high $\beta$ values ($3 \times 10^{-6}$--$1 \times 10^{-6}$), where strong regularisation compresses the adapter to a very small effective rank without sacrificing predictive performance. This is consistent with the motivation in Section~\ref{sec:intro}: in low-data regimes, the KL penalty acts as a beneficial inductive bias, favouring compact representations that generalise better. On larger datasets (MNLI, QQP), accuracy remains stable across low $\beta$ values but falls sharply once $\beta$ is large enough to substantially reduce rank, indicating that these tasks extract more value from higher effective capacity than aggressive regularisation.

Together with the $\tau$ sweep in Appendix~\ref{app:tau_ablation}, these results suggest that $\beta$ and $\tau$ provide complementary, interpretable control over the accuracy--compression frontier, with dataset size offering intuitive guidance on where to operate.

\section{Reasoning Experiment Hyperparameters}
\label{app:reasoning_bench}

We evaluate BayesLoRA against comparable uncertainty-aware baselines on
\textsc{ARC-Challenge}, \textsc{ARC-Easy}, \textsc{WinoGrande-S},
\textsc{WinoGrande-M}, \textsc{BoolQ}, and \textsc{OpenBookQA} using a
Llama~2~7B backbone. All results are averaged over three random seeds. Table~\ref{tab:reasoning_hparams} lists the full training, system, and uncertainty-specific hyperparameters used across all reasoning tasks.

\begin{table}[H]
\centering
\caption{Training, system, and uncertainty-specific parameters for reasoning benchmark experiments.}
\label{tab:reasoning_hparams}
\vspace{4pt}
\begin{tabular}{ll}
\toprule
\textbf{Parameter} & \textbf{Value} \\
\midrule
Learning rate & $1\times10^{-4}$ \\
Max steps & 5000 \\
Batch size & 2 \\
Gradient accumulation steps & 1 \\
\midrule
LoRA rank & 8 \\
LoRA alpha & 16 \\
LoRA dropout & 0.1 \\
LoRA target modules & \texttt{[q, v, lm\_head]} \\
\midrule
Evaluation batch size & 2 \\
Evaluation metrics & Accuracy, F1 \\
Evaluation frequency & Every 100 steps \\
Number of MC passes ($k$) & \texttt{[0,5,10]} \\
\midrule
KL scale & $1\times10^{-7}$ \\
Log-$\alpha$ threshold & 4.0 \\
Initial log $\sigma$ & $-8.0$ \\
Pruning steps & \texttt{[1000, 2000, 3000]} \\
Tie alpha per rank & True \\
Local reparameterisation & True \\
\bottomrule
\end{tabular}
\end{table}

\subsection{Inference-Time Monte Carlo Sample Sweep}
\label{app:mc_sample_sweep}

\begin{figure*}[t]
    \centering
    \includegraphics[width=0.82\textwidth]{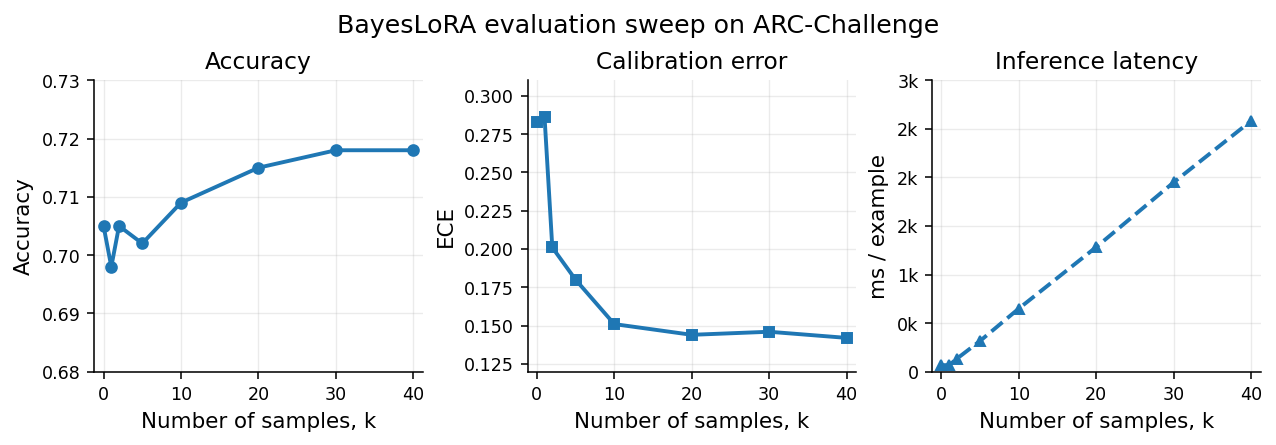}
    \caption{
    \textbf{Inference-time Monte Carlo sample sweep on \textsc{ARC-Challenge}.}
    Accuracy improves modestly and saturates as the number of posterior samples $k$ increases,
    while calibration error decreases substantially. Inference latency grows approximately
    linearly with $k$, reflecting the expected cost of Monte Carlo marginalisation.
    }
    \label{fig:arc_sample_sweep}
\end{figure*}

Figure~\ref{fig:arc_sample_sweep} shows the effect of the number of Monte Carlo samples at inference time on \textsc{ARC-Challenge}. Accuracy improves only modestly and saturates quickly, while calibration error decreases substantially with additional samples. In contrast, inference latency grows approximately linearly with $k$, exposing a clear accuracy--calibration--latency trade-off. This motivates reporting $k \in \{0,5,10\}$ in Table~\ref{tab:reasoning_main}, where $k=0$ corresponds to deterministic posterior-mean inference and $k>0$ uses Monte Carlo marginalisation over the learned rank-space posterior.

\end{document}